\def\paperID{13729}
\def\confName{ICCV}
\def\confYear{2025}

\def\paperTitle{MixRI: Mixing Features of Reference Images for Novel Object Pose Estimation}

\def\authorBlock{
    Xinhang Liu$^{1,2}$ \qquad
    Jiawei Shi$^{1,2}$ \qquad
    Zheng Dang$^{3}$ \quad
    Yuchao Dai$^{1,2}$ \thanks{\ Yuchao Dai is the corresponding author. \\ This work was supported in part by the National Natural Science Foundation of China under Grants 62271410 and 12150007.} \\
    School of Electronics and Information, 
    Northwestern Polytechnical University$^{1}$ \\  Shaanxi Key Laboratory of Information Acquisition and Processing$^{2}$\\ CVLab, EPFL, Switzerland$^{3}$ \\

    {\tt\small \{xinhangliu, sjw2018\}@mail.nwpu.edu.cn, zheng.dang@epfl.ch, daiyuchao@nwpu.edu.cn}
}

\newif\ifreview 
\newif\ifarxiv \newcommand{\arxiv}{\arxivtrue}
\newif\ifcamera 
\newif\ifrebuttal 

\arxiv 

\pdfoutput=1
\documentclass[10pt,twocolumn,letterpaper]{article}
\ifreview \usepackage[review]{cvpr} \fi
\ifarxiv \usepackage[pagenumbers]{cvpr} \fi
\ifrebuttal \usepackage[rebuttal]{cvpr} \fi
\ifcamera \usepackage{cvpr} \fi

\usepackage{graphicx}	
\usepackage{amsmath}	
\usepackage{amssymb}	
\usepackage{booktabs}
\usepackage{times}
\usepackage{microtype}
\usepackage{epsfig}
\usepackage{caption}
\usepackage{float}
\usepackage{placeins}
\usepackage{color, colortbl}
\usepackage{stfloats}
\usepackage{enumitem}
\usepackage{tabularx}
\usepackage{xstring}
\usepackage{multirow}
\usepackage{xspace}
\usepackage{url}
\usepackage{subcaption}
\usepackage{xcolor}
\usepackage[hang,flushmargin]{footmisc}

\usepackage[normalem]{ulem}
\usepackage{pifont}
\usepackage{enumitem}

\ifcamera \usepackage[accsupp]{axessibility} \fi

% ...

\ifarxiv  \fi

\newcommand{\R}[1]{{%
    \textbf{%
        \ifstrequal{#1}{1}{\textcolor{red}{R#1}}{%
        \ifstrequal{#1}{2}{\textcolor{blue}{R#1}}{%
        \ifstrequal{#1}{3}{\textcolor{magenta}{R#1}}{%
        \ifstrequal{#1}{4}{\textcolor{teal}{R#1}}{%
                           \textcolor{cyan}{R#1}%
        }}}}%
    }%
}}

\usepackage{xr-hyper}

\makeatletter
\newcommand*{\addFileDependency}[1]{
  \typeout{(#1)}
  \@addtofilelist{#1}
  \IfFileExists{#1}{}{\typeout{No file #1.}}
}

\makeatother
\newcommand*{\myexternaldocument}[1]{
    \externaldocument{#1}
    \addFileDependency{#1.tex}
    \addFileDependency{#1.aux}
}

\definecolor{cvprblue}{rgb}{0.21,0.49,0.74}
\usepackage[pagebackref,breaklinks,colorlinks,allcolors=cvprblue]{hyperref}
\usepackage[capitalize]{cleveref}
\crefname{section}{Sec.}{Secs.}
\crefname{table}{Table}{Tables}
\crefname{figure}{Fig.}{Figs.}

\ifarxiv \crefname{appendix}{App.}{Apps.}
\else \crefname{appendix}{Suppl.}{Suppls.} \fi

\unless\ifarxiv \myexternaldocument{_supplementary} \fi

\begin{document}
\title{\paperTitle}
\author{\authorBlock}
\maketitle

\begin{abstract}
We present MixRI, a lightweight network that solves the CAD-based novel object pose estimation problem in RGB images. It can be instantly applied to a novel object at test time without finetuning. We design our network to meet the demands of real-world applications, emphasizing reduced memory requirements and fast inference time. Unlike existing works that utilize many reference images and have large network parameters, we directly match points based on the multi-view information between the query and reference images with a lightweight network. Thanks to our reference image fusion strategy, we significantly decrease the number of reference images, thus decreasing the time needed to process these images and the memory required to store them. Furthermore, with our lightweight network, our method requires less inference time. Though with fewer reference images, experiments on seven core datasets in the BOP challenge show that our method achieves comparable results with other methods that require more reference images and larger network parameters\footnote{project page: \url{https://npucvr.github.io/MixRI/}}.
\end{abstract}
\vspace{-1.5em}
\section{Introduction}
\label{sec:intro}

Six degrees-of-freedom (6DoF) object pose estimation predicts the orientation and location of a target object in 3D space. This is crucial for embodied AI, as intelligent agents must comprehend and interact with their environments to perform tasks such as robotic manipulation and augmented reality applications \cite{marchand2015pose,deng2020self}. In recent years, 6DoF object pose estimation accuracy has been significantly improved with deep learning~\cite{xiang2017posecnn,kehl2017ssd,labbe2020cosypose,li2018deepim,peng2019pvnet,zakharov2019dpod,su2022zebrapose,rad2017bb8,xu2022rnnpose,he2021ffb6d,wang2021gdr,lin2024hipose,li2019cdpn,bukschat2020efficientpose}. However, these methods require the same object during training and test time and rely on generating abundant synthetic data for each object during the training stage, which is tedious to deploy in practical situations. To satisfy the practical need in industrial and daily life, CAD-based novel object pose estimation (During the training stage, the specific object to be inferred at test time is unknown.) was introduced~\cite{hodan2023bop} and more and more methods~\cite{labbe2022megapose, shugurov2022osop, nguyen2024gigaPose, ausserlechner2023zs6d,nguyen2022templates,chen2023zeropose,pitteri20203d,huang2024matchu,caraffa2024freeze,zhao2022fusing,foundationposewen2024,wang2024object,ornek2023foundpose, moon2024genflow} arise. 
\begin{figure}[t]
    \centering
    \includegraphics[width=0.5\textwidth]{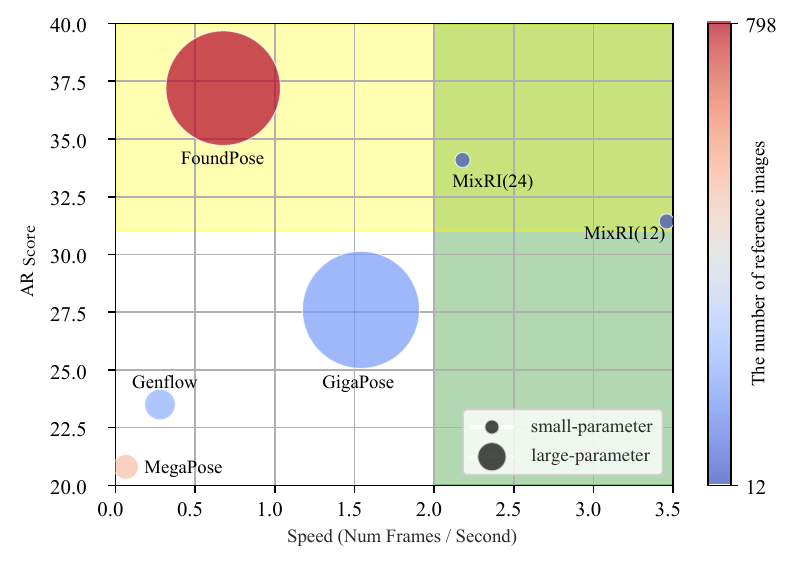}
    \caption{\textbf{Comparison of different methods.} The area of each bubble is proportional to the size of the network parameters, and the color indicates the number of reference images used. The detection\cite{nguyen2023cnos} stage was removed in all speed evaluations. MixRI achieves competitive results while using fewer reference images, a smaller network, and providing shorter inference time.}
    \label{fig:scatter_teaser}
    \vspace{-1.5em}
\end{figure}
\begin{figure*}[ht]
    \centering
    \vspace{-1em}
    \includegraphics[scale=0.55]{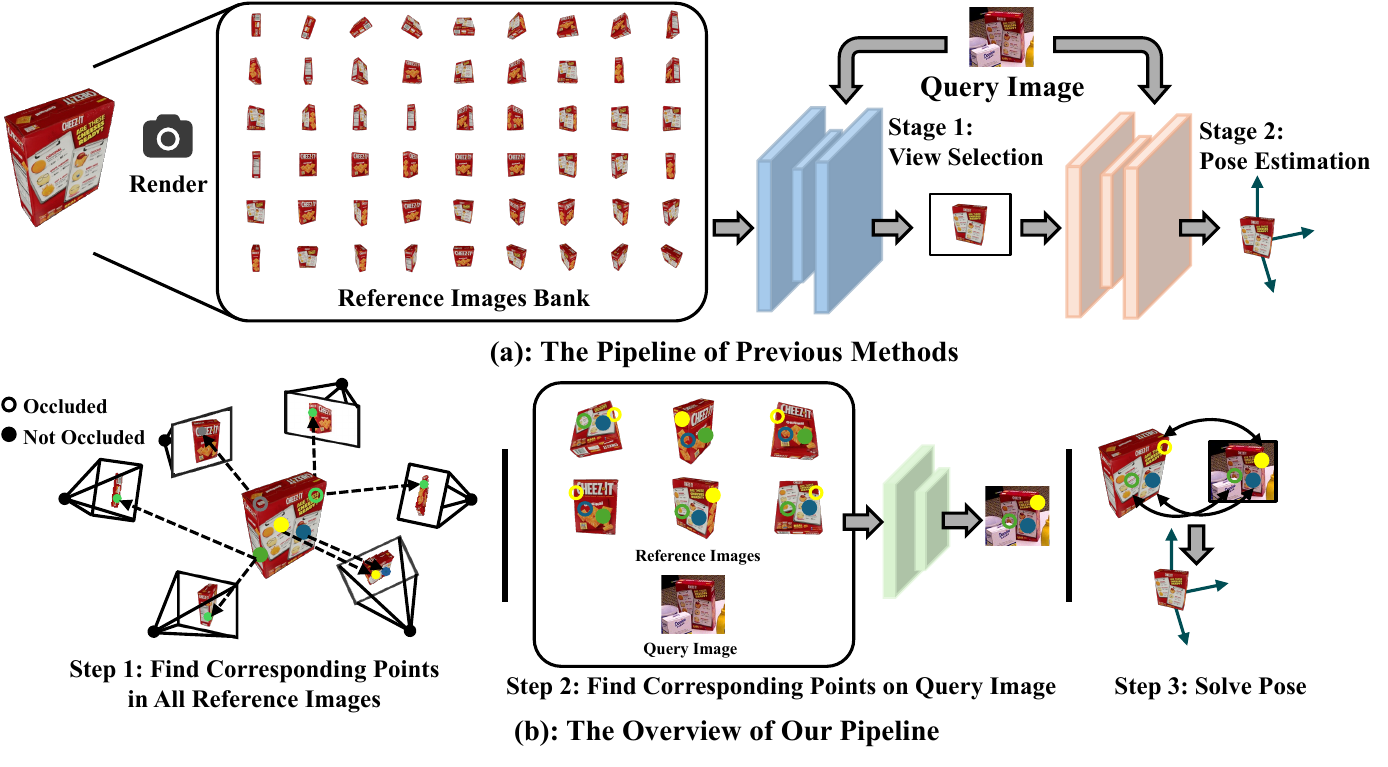}
    \vspace{-1em}
    \caption{\textbf{Comparison between our pipeline and previous methods}. Unlike existing two-stage methods~\cite{nguyen2024gigaPose,labbe2022megapose,ausserlechner2023zs6d,ornek2023foundpose} that first retrieve the closest reference image from abundant reference images, we directly predict the location of sampled 3D object points on the query image from their projections on all reference images. It constructs the 2D-3D correspondence and can solve the 6DoF pose. Hollow points indicate invisibility, while solid points indicate visibility.}
    \label{fig:pipeline}
    \vspace{-1.5em}
\end{figure*}

To enable methods to adapt to different objects without retraining, recent works utilize reference images. These images are typically rendered offline and stored in memory~\cite{caraffa2024freeze, ausserlechner2023zs6d, chen2023zeropose, ornek2023foundpose, nguyen2024gigaPose, labbe2022megapose, moon2024genflow}. For edge AI and efficient deployment on edge devices in practical embodied AI applications, a compact memory cache for both reference images and network parameters is essential, along with the need for fast inference time \cite{li2019edge}. However, as shown in \Cref{fig:scatter_teaser}, although recent works have successfully pursued high pose estimation accuracy, they do so at the cost of requiring extensive reference images, large network parameters, and slow inference time. As shown in \Cref{fig:pipeline} (a), existing methods~\cite{ausserlechner2023zs6d,labbe2022megapose,nguyen2024gigaPose,ornek2023foundpose,shugurov2022osop, moon2024genflow} usually divide the pose estimation into two stages: a view selection stage to retrieve the closest reference image and a pose estimation stage to estimate the 6DoF pose of the query image by comparing it with the selected reference image. As the number of reference images for each object increases, this restricts the algorithm's ability to scale to a larger number of objects, as each object requires rendering a substantial number of reference images to be stored in memory \cite{ornek2023foundpose}. Besides, these works mainly rely on networks with a large number of parameters. The large number of rendered reference images and extensive network parameters present a significant challenge for devices with limited computing power and memory. Additionally, some previous works \cite{ornek2023foundpose,nguyen2024gigaPose} require the pre-extraction of image features, which are stored in memory as well. These limitations restrict their practical application in edge AI and limit the number of unseen objects whose poses can be estimated, as algorithms typically run in real time on edge devices with limited memory.

In this paper, we aim to solve the novel pose estimation problem using a lightweight network with a minimal number of reference images, designed to support edge AI \cite{li2019edge}. Additionally, our approach eliminates the need for pre-extraction of features. Inspired by previous work that utilizes matching to compute the pose~\cite{nguyen2024gigaPose,ausserlechner2023zs6d,ornek2023foundpose}, we also try to build our method in a matching framework. However, when the number of reference images is significantly reduced, the closest reference image may still suffer from wide rotation with the query image, known as a wide baseline in matching, which is challenging to address \cite{jin2021image}. To effectively solve the matching problem, we design our method from the following observation: 1) Given that multi-view geometry can provide more information~\cite{hartley2003multiple}, we can use multi-view information to enhance the matching procedures. 2) When the reference images are few, the views become sparse, and occlusion becomes more common. This makes matching more necessary to integrate multi-view information and handle occlusion scenarios. Based on the above two observations, we design our network by aggregating all the reference image features and propose \textbf{MixRI} (\textbf{Mix} \textbf{R}eference \textbf{I}mages), which is a lightweight network and requires only few reference images as input to solve the novel object pose estimation.

As shown in~\Cref{fig:pipeline} (b), we gather all the 2D information belonging to the same 3D object points in all reference images and mix their features. We train a lightweight network to fully fuse the feature of all reference images and enhance the query image feature with the reference images, which makes the matching more stable and accurate. We also design the occlusion detection as part of the network's output, making our network handle the occlusion automatically. In contrast to the usual feature matching problem \cite{sun2021loftr, detone2018superpoint, xie2024deepmatcher, xie2021cotr} between paired images, our method requires establishing correspondences from multiple reference images to a single query image. Specifically, the objective is to determine the position and occlusion flag of each 3D object point on the query image, based on its projections on all the reference images. Upon establishing these correspondences, the corresponding 2D points for the 3D object points are then derived. With a sufficient number of 2D-3D correspondences, the object pose can subsequently be computed using the PnP algorithm~\cite{terzakis2020consistently} within the RANSAC~\cite{fischler1981random} framework.

Our experiments demonstrate that, under the setting without refinement, our method achieves comparable results despite utilizing 33$\times$ fewer reference images than the most accurate method \cite{ornek2023foundpose} and being 2$\times$ faster than the fastest method \cite{nguyen2024gigaPose}. This suggests that a heavy reliance on a large number of reference images and large network parameters may not be the only way for achieving high performance.

In summary, our contributions can be summarized as:

\begin{itemize}
    \item We present MixRI, a lightweight network for RGB-based novel object pose estimation. It requires only 12 reference images and does not necessitate the offline pre-extraction of image features.
    \item We propose a View-Aggregated Point Matching module that can find correspondences on the query image based on multiple reference images simultaneously and give occlusion prediction.  
    \item MixRI achieves comparable performance to state-of-the-art approaches while using significantly fewer reference images, significantly fewer network parameters, shorter preparation time for reference images, and faster inference speed, making it more suitable for practical applications.
\end{itemize}

\section{Related Work}
\label{sec:related}

\textbf{Pose Estimation of Known Objects or Categories.}
6DoF object pose estimation has been widely studied for decades as a fundamental vision problem~\cite{lowe1987three,lowe1999object}, solved from the traditional methods~\cite{lowe1987three, collet2011moped, lowe1999object,drost2010model} to deep learning methods~\cite{xiang2017posecnn,su2022zebrapose,rad2017bb8,zakharov2019dpod,li2019cdpn,peng2019pvnet,wen2020robust,he2021ffb6d,lin2024hipose}. Early deep learning-based methods for pose estimation focused on the seen object pose estimation problem either based on template matching~\cite{kehl2017ssd} or feature matching followed by PnP~\cite{chen2022epro,bukschat2020efficientpose} for 2D-3D correspondences~\cite{peng2019pvnet,rad2017bb8,zakharov2019dpod,su2022zebrapose,bukschat2020efficientpose,li2019cdpn} or least squares fitting for 3D-3D correspondences~\cite{he2021ffb6d, wang2021gdr,he2020pvn3d, chen2020learning, lin2024hipose}. To compensate for the expensive retraining required for a new object in seen object scenario, one approach involves making the trained network generalize to known categories~\cite{manhardt2020cps++,wang2019normalized,lin2022single,tian2020shape,li2024category,chen2024secondpose,lee2023tta}. While these approaches can estimate unseen objects, they assume all objects belong to the same category, which still limits their application. In contrast, our work can be generalized to arbitrary objects.

\noindent\textbf{Pose Estimation of Unseen Categories.}
Recently, some works have focused on novel object pose estimation, where the target category is not seen during the training stage. This can be addressed when an object model is available~\cite{nguyen2024gigaPose,chen2023zeropose,pitteri20203d,huang2024matchu,caraffa2024freeze,shugurov2022osop,nguyen2024gigaPose,ausserlechner2023zs6d,zhao2022fusing,nguyen2022templates,labbe2022megapose,foundationposewen2024,wang2024object,ornek2023foundpose,moon2024genflow} or with reference images~\cite{sun2022onepose, he2022onepose++,wen2023bundlesdf,foundationposewen2024,liu2022gen6d,pan2023learning,he2022fs6d,cai2024gs}. These works either use RGB images~\cite{labbe2022megapose,nguyen2024gigaPose,ausserlechner2023zs6d,liu2022gen6d,pan2023learning,sun2022onepose,he2022onepose++,zhao2022locposenet,wang2024object,shugurov2022osop,moon2024genflow} or RGB-D images~\cite{labbe2022megapose, foundationposewen2024,lin2023sam,huang2024matchu,shugurov2022osop,okorn2021zephyr,chen2023zeropose,caraffa2024freeze,moon2024genflow} as input. For scenes with available object models, methods can be divided into feature-matching methods~\cite{ausserlechner2023zs6d, nguyen2024gigaPose,chen2023zeropose,pitteri20203d,huang2024matchu,caraffa2024freeze,ornek2023foundpose} and template matching methods~\cite{shugurov2022osop,nguyen2024gigaPose,ausserlechner2023zs6d,zhao2022fusing,nguyen2022templates,labbe2022megapose,foundationposewen2024,wang2024object,ornek2023foundpose, moon2024genflow}. However, for those feature-matching methods, they primarily use a retrieval-based strategy, which actually applies a template matching method to get the closest reference image and build the correspondence between the closest reference image and the query image~\cite{ausserlechner2023zs6d,nguyen2024gigaPose,chen2023zeropose,moon2024genflow,ornek2023foundpose}. Unlike those works, our method is pure feature-matching. We directly build the correspondences across all the reference images and the query image, which significantly reduce the number of reference images needed. This reduces the time needed to render reference images and avoids caching pre-computed features for view selection, saving substantial memory.

\noindent\textbf{Local Feature Matching \& Point Tracking.}
Local feature matching establishes precise correspondences between images, typically in paired image scenarios. Early works use a detector-based approach first employing some well-established handcrafted features such as SIFT~\cite{lowe2004distinctive} and ORB~\cite{rublee2011orb} followed by a feature matching algorithm like nearest-neighbor searches. With the help of deep learning, more complex methods~\cite{detone2018superpoint,mishchuk2017working,revaud2019r2d2,dusmanu2019d2,liu2019gift} arise. Recently, more works~\cite{sun2021loftr,xie2021cotr,xie2024deepmatcher,chen2022aspanformer} are detector-free methods, and they can efficiently handle extreme circumstances such as texture-less regions or substantial viewpoint changes. Our work focuses on feature matching, but we build the correspondences between the query image and multiple reference images of the same 3D object points, which differs from the matching between two images. Those methods struggle to handle occlusion and massive changes in viewpoint, which can not predict the occlusion information as well.

Besides, some works focus on tracking points across continuous frames~\cite{doersch2022tap,dino_tracker_2024,wang2023tracking,harley2022particle,luo2024continuous}. In those works, adjacent frames, along with the initial points to be tracked, are input into the network. The network will then output the points' trajectories across the sequence, along with occlusion information. As with works in point tracking, we also build the correspondence along all the images and output the occlusion information. However, in our method, the reference images are unordered and not consecutive. Furthermore, we have the locations of the query 3D object points on all reference images, which are the projections of the same 3D object points onto each reference image, and some can be occluded. While in point tracking work, query points usually belong to a single frame and must be visible.

\section{Method}
\label{sec:method}

Our method is built on finding the correspondences between the 2D pixel locations and 3D object point locations, followed by a RANSAC-based PnP framework to compute the 6DoF pose. We break down the pose estimation problem to a matching problem. However, different from the previous matching-based pose estimation, we neither directly find 2D-3D correspondences between the query image and 3D object points~\cite{peng2019pvnet,zakharov2019dpod, sun2022onepose,he2022onepose++} nor find 2D-2D correspondences between paired images as is done in classic local feature matching~\cite{sun2021loftr,xie2021cotr,ausserlechner2023zs6d,nguyen2024gigaPose}. Instead, we focus on finding correspondences between multiple reference images and the query image. Given multiple projections of one 3D object point, some of which may be occluded, we aim to find its projection coordinate and occlusion flag on the query image.

\begin{figure*}[ht]
    \centering
    \includegraphics[scale=0.45]{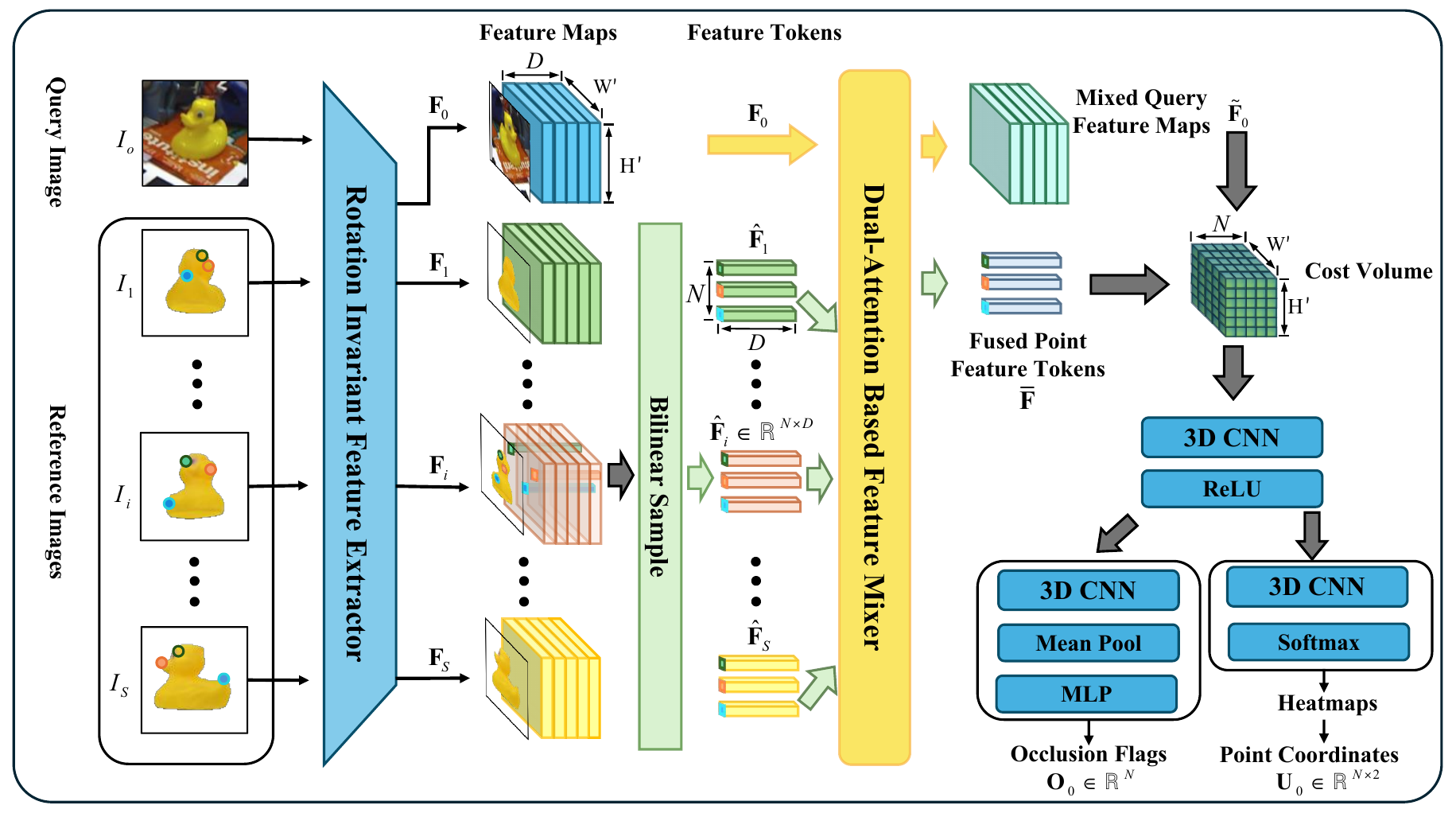}
    \setlength{\belowcaptionskip}{-10pt}
    \vspace{-0.5em}
    \caption{\textbf{Overview of our network}. Unlike existing methods~\cite{labbe2022megapose,ausserlechner2023zs6d,nguyen2024gigaPose,ornek2023foundpose}, which use two stages to compute the pose, we directly input \textbf{all} reference images into our network without a view selection stage. After obtaining all features of the projection belonging to one 3D object point $\mathbf{p}_k$, we use Dual-Attention Based Feature Mixer to fuse their features with the query image feature. Then, we build the cost volume followed by two separate heads to predicate the projection of $\mathbf{p}_k$ on the query image, including the occlusion flag as well as the coordinate.}
    \label{fig:overview}
    \vspace{-0.5em}
\end{figure*}
\subsection{Preliminary}
Our task is CAD-based novel object pose estimation with RGB images, where the primary goal is to estimate the pose $\mathbf{T}_0$ of the query image $I_0$. It is worth noting that the objects observed in $I_0$ used for testing are unseen in the training dataset,~\ie, $\mathcal{O}_{train}\cap\mathcal{O}_{test}=\emptyset$~\cite{zhao2022locposenet}.
After detecting and segmenting the object in the query image, we can estimate the pose of the novel object using a series of reference images and depths rendered by the corresponding object model.
In other words, given $S$ reference images $\{I_{1}, I_{2}, \dots, I_{S}\}$ showing the same object under various viewpoints, for which the object pose $\mathbf{T}_i\!=\!(\mathbf{R}_i, \mathbf{t}_i) \in SE(3)$, intrinsic matrix $\mathbf{K}_{i}\in\mathbb{R}^{3\times 3}$ and depth $\mathbf{D}_{i}$ are known, we focus on estimating the novel object pose in the query image.

\subsection{Correspondences between Reference Images}
\label{sec::gt}
Before training and testing, to find the projections of each 3D object point across all reference images, we first sample on each reference image $i$ to obtain 2D pixel coordinates $\mathbf{u}_{i,k}\in \mathbb{R}^{2\times 1}$. Using the given ground truth pose $\mathbf{T}_{i}$ and depth $d_{i,k}$, we can then recover the corresponding 3D object point in the object model coordinate:
\begin{equation}
    \mathbf{p}_{k} = d_{i, k}\mathbf{T}_{i}^{-1}\mathbf{K}_{i}^{-1}\tilde{\mathbf{u}}_{i, k},
\end{equation}
where, $\tilde{\mathbf{u}}_{i, k}$ is homogeneous coordinate, $1 \leq k \leq N$ and $N$ is the total number of sampled points. Besides, we can calculate new 2D coordinates on other reference image $j$ by projecting the 3D object points: 
\begin{equation}
    \tilde{\mathbf{u}}_{j, k} = \frac{1}{\tilde{d}_{j, k}}\mathbf{K}_{j}\mathbf{T}_{j}\mathbf{p}_{k},
\end{equation}
where, $\tilde{d}_{j, k}$ is the depth of point $\mathbf{p}_{k}$ transformed to the $j$-th camera coordinate. In this manner, we can obtain the 2D pixel coordinates $\{\mathbf{u}_{i,k}\}$ of the $N$ sampled points on $S$ reference images and their 3D object coordinates $\{\mathbf{p}_{k}\}$.
However, since each point may be occluded by the object itself or other objects, we set the occlusion flag $O_{i,k}\!\in\!\{0,1\}$ for each $\mathbf{u}_{i,k}$. Similar to the Z-buffer algorithm \cite{catmull1978hidden}, we mark points that are projected outside the image or occluded, \ie, $\left |\tilde{d}_{j, k} -{d}_{j, k} \right | > \tau$, as occlusion~($O_{j,k}=1$).

Next, we aim to compute the corresponding point on the query image. Supposing among $S$ reference images, for a given 3D object point $\mathbf{p}_k$, $S_o$ of all its projections are occluded and $S_v$ are visible (where $S_v+S_o=S$), one approach is to use $S_v$ reference images to match pairwise with it, and finally merge all the matching results to get $\mathbf{u}_{0,k}$, which is the projection of $\mathbf{p}_k$ on the query image $I_0$. However, this approach can not handle the situation when $\mathbf{u}_{0,k}$ is occluded and fails to fully utilize the multi-view information. To address this, for a 3D object points $\mathbf{p}_k$, we develop a network to directly fuse all the projections' information around $\mathbf{u}_{i, k}, 1\leq i \leq S$ in the feature space and locate the corresponding 2D pixel coordinate $\mathbf{u}_{0, k}$ on the query image, as shown in~\Cref{sec::architecture}.

\begin{figure*}[ht]
    \centering
    \includegraphics[scale=0.43]{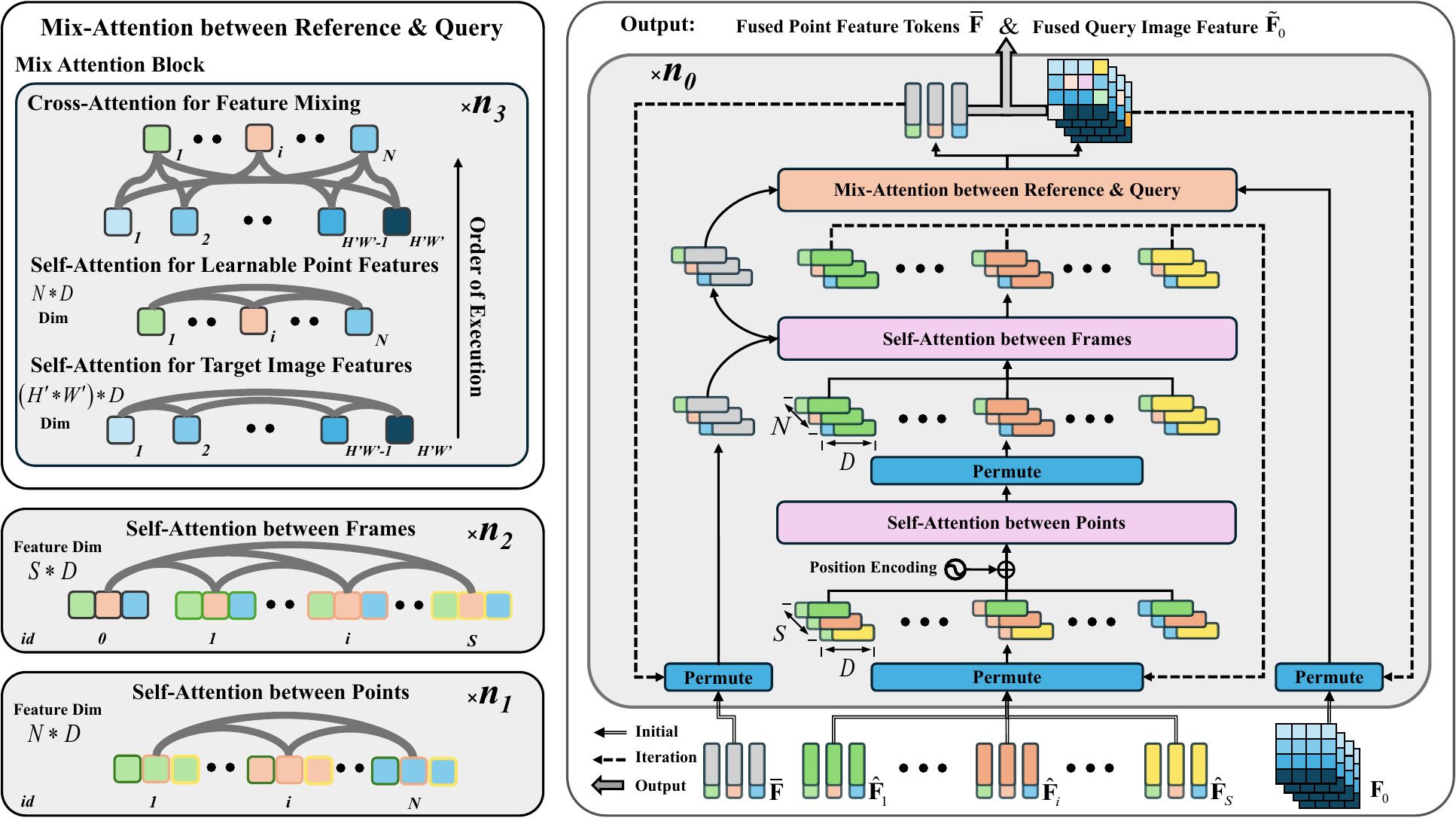}
    \vspace{-0.5em}
    \caption{\textbf{Overview of the Dual-Attention Based Feature Mixer.} The mixer consists of three modules: SAP (Self-Attention between Points) , SAF (Self-Attention between Frames), and MARQ (Mix-Attention between Reference \& Query). These modules perform attention operations between N points within one reference image, between S frames, and between reference images and the query image, respectively.}
    \label{fig:figure3}
    \vspace{-1.2em}
\end{figure*}

\subsection{View-Aggregated Point Matching}
\label{sec::architecture}
\Cref{fig:overview} shows an overview of our network architecture. The query image, along with the reference images, are sent to the encoder $f_{enc}$ together. $f_{enc}$ is a ResNet-like backbone~\cite{he2016deep} sharing the same weights between the query and reference images.
Since objects in images may have various poses and the features of the matching points are independent of the poses, we utilize~\cite{e2cnn}
in the encoder to ensure that the extracted features are rotation-invariant. After feature extraction, we obtain $S+1$ feature maps:
\begin{equation}
    \mathbf{F}_{i} = f_{enc}(I_{i}) \in \mathbb{R}^{H'\times W' \times D}, 0\leq i \leq S,
\end{equation}
where $H'\!=\!H/8,W'\!=\!W/8$. With pre-sampled 2D projections $\{\mathbf{u}_{i, k}\}_{i=1}^{S}$ and occlusion flags $\{O_{i,k}\}_{i=1}^{S}$ on reference images, we retrieve the feature tokens $\{\hat{\mathbf{F}}_{i}\in \mathbb{R}^{N\times D}\}_{i=1}^{S}$ using bilinear interpolation.
Next, for all 3D object points $\{\mathbf{p}_k\}_{k=1}^{N}$, we gather all feature tokens in the feature space and fuse them with the query image feature $\mathbf{F}_0$ using attention mechanism (see \Cref{attention} for details). We denote the final gathered feature and fused query image feature as $\bar{\mathbf{F}} \in \mathbb{R}^{N\times D}$, ${\tilde{\mathbf{F}}}_{0} \in \mathbb{R}^{H'\times W'\times D}$, respectively. 
Then we can build the cost volume $\mathbf{C} \in \mathbb{R}^{H'\times W'\times N}$ with the query image feature ${\tilde{\mathbf{F}}}_{0}$.

For the given cost volume $\mathbf{C}$, we first extract the information with a Conv3D backbone. After that, we use two separate heads to predict the corresponding 2D heatmap of the 3D object points and the associated occlusion flags. To convert the heatmap to valid 2D coordinates, we use ``spatial soft argmax''~\cite{doersch2022tap}, which computes the argmax of the heatmap and then estimates the spatial average position within a radius around the argmax location. Lastly, we calculate the 6DoF pose directly using RANSAC-based SQ-PnP~\cite{terzakis2020consistently} algorithm. Note that we only use the 2D location where the predicted occlusion flag is \textit{False}, \ie, where $O_{0,k}\le\tau_\text{occ}$.

\subsection{Dual-Attention Based Feature Mixer}
\label{attention}
 Previous methods \cite{ausserlechner2023zs6d,nguyen2024gigaPose,ornek2023foundpose} select the reference image closest to the query image for matching. Unlike them, we fuse the feature tokens extracted from all reference images to obtain the final tokens that contain more viewpoint information. We design the Dual-Attention module, an overview of which is illustrated in~\Cref{fig:figure3}. Referring to~\cite{dosovitskiy2020vit}, we first initialize the fused feature tokens $\bar{\mathbf{F}}$ as learnable parameters, then input the feature tokens $\hat{\mathbf{F}} \in \mathbb{R}^{N\times S \times D}$ of all reference images, occlusion flags  $\mathbf{O}\in\mathbb{R}^{S\times N}$ of these tokens, and the feature map $\mathbf{F}_{0}$ of the query image. 

Firstly, in order to integrate the spatial location information of $N$ points, we permute $\hat{\mathbf{F}}$ as $\hat{\mathbf{F}}^{N}=g_{N}(\hat{\mathbf{F}})\in \mathbb{R}^{S \times N\times D}$. Then we input $\hat{\mathbf{F}}^{N}$ into the SAP (\textbf{S}elf-\textbf{A}ttention between \textbf{P}oints) module for spatial dimension mixing after adding position encoding~\cite{sun2021loftr}. Secondly, the features mixed on $N$-dim are permuted into frame-first form, \ie, $\hat{\mathbf{F}}^{S}=g_{S}(\hat{\mathbf{F}})\in \mathbb{R}^{N\times S\times D}$, and then sent to SAF (\textbf{S}elf-\textbf{A}ttention between \textbf{F}rames) module to perform Self-Attention calculation with learnable fused tokens $\bar{\mathbf{F}}$ on $S$-dim. These permutations enables the first dimension to be merged with the batch size dimension, thereby accelerating computation when multiple objects are predicted concurrently. Finally, the adjusted tokens $\hat{\mathbf{F}}$ are input for the next iteration. To further augment the features, we fuse $\bar{\mathbf{F}}$ with the permuted query image feature $\mathbf{F}_{0}\in \mathbb{R}^{(H'W') \times D}$ using MARQ (\textbf{M}ix-\textbf{A}ttention between \textbf{R}eference \& \textbf{Q}uery) module, which consists of two Self-Attention and two Cross-Attention layers. It performs the Self-Attention separately on the query image feature $\mathbf{F}_0$ and the learnable fused token $\bar{\mathbf{F}}$, and Cross-Attention between the two~\cite{sun2021loftr}. The Dual-Attention module can be written as:
\begin{equation}\label{eq:attention}
    \begin{aligned}
    \left\{\begin{array}{l}
\hat{\mathbf{F}}=\mathrm{Self_N}(g_{N}(\hat{\mathbf{F}}),~\mathbf{O}),~for~n_{1}~iters\\
\hat{\mathbf{F}},~\bar{\mathbf{F}}=\mathrm{Self_{S}}(g_{S}(\hat{\mathbf{F}}),~\bar{\mathbf{F}},~\mathbf{O}),~for\ n_{2}\ iters\\
\bar{\mathbf{F}},~{\mathbf{F}}_{0}=\mathrm{Mix}(\bar{\mathbf{F}},~{\mathbf{F}}_{0},~\mathbf{O}), ~for\ n_{3}\ iters
\end{array}\right.
    \end{aligned},
\end{equation}
where $\mathrm{Self}_{N},~\mathrm{Self}_{S},~\mathrm{Mix}$ denote Self-Attention on $N$-dim, Self-Attention on $S$-dim, and Mix-Attention, respectively. Mask $\mathbf{O}$ is used in masked attention mechanism,  introduced to correct the weight matrices of attention, which avoids fusing erroneous features corresponding to occluded points. Iterate~\Cref{eq:attention} for $n_{0}$ times and take the obtained $\bar{\mathbf{F}}$ and ${\tilde{\mathbf{F}}}_{0}$ as the final fused results. By combining Self/Cross-Attention, Self-Attention on $N$/$S$-dim, our Dual-Attention module can automatically perform fusion based on the degree of similarity between point-to-point, frame-to-frame and reference-to-query during feature mixing. This module is crucial to accomplishing the view-aggregated point matching task, as we will show in~\Cref{sec:ablation}.

\subsection{Training Losses}
The total loss for each 3D object point consists of two components: occlusion supervision and location supervision. The occlusion loss is supervised using a BCE loss:

\begin{equation}
    L_{\text{occ}} = \text{BCE}(O_{gt,k}, O_{0,k}),
\end{equation}
where $O_{0,k}$ is network output for the $k^{th}$ 3D object point and $O_{gt,k}$ is the corresponding ground truth occlusion flag.

For location supervision, we use the Huber loss to regress the 2D coordinates of the projection of the 3D object points when the occlusion flag is False:
\begin{equation}
    L_{\text{loc}} = \text{Huber}(\mathbf{U}_{gt,k}, \mathbf{U}_{0,k})\cdot 1\{O_{gt,k}=0\},
\end{equation}
where $\mathbf{U}_{gt,k}$ and $\mathbf{U}_{0,k}$ is the ground truth 2D coordinate and predicted 2D coordinate for $k^{th}$ 3D object point. $1\{\cdot\}$ is the indicator function. The total loss is a weighted sum of the occlusion and location losses:
\begin{equation}
    L = L_\text{occ} + \lambda L_\text{loc},
\end{equation}
with $\lambda=100$. The final loss is the mean of the losses across $N$ sampled points.


\section{Experiments}
\label{sec:experiments}

In this section, we first present the experiment setup. Then, we compare our method with the state-of-the-art methods for novel object pose estimation on seven core datasets of the BOP challenge~\cite{hodan2023bop}. We also provide further ablation studies to evaluate our method.

\subsection{Datasets}
We train our network entirely with synthetic images from the GSO-Dataset \cite{labbe2022megapose}. Following previous work \cite{nguyen2024gigaPose,labbe2022megapose,ornek2023foundpose,moon2024genflow}, we evaluate our method on seven core BOP datasets~\cite{hodan2023bop}, including LM-O~\cite{brachmann2014learning}, YCB-V~\cite{xiang2017posecnn}, T-LESS~\cite{hodan2017tless}, TUD-L~\cite{hodan2018bop}, IC-BIN~\cite{doumanoglou2016recovering}, HomebrewedDB(HB)~\cite{kaskman2019homebreweddb} and ITODD~\cite{drost2017introducing}. Detailed information is explained in the supplementary material. We follow the official test splits of each dataset and use the off-the-shelf object detector CNOS~\cite{nguyen2023cnos}. In practical situations, this detector can be replaced by any other lightweight tracking method or an object-specific segmentation network.

\subsection{Evaluation Metrics}
For all experiments, we use the standard evaluation protocol in the BOP challenge~\cite{hodan2018bop}, including Visible Surface Discrepancy (VSD), Maximum Symmetry-Aware Surface Distance (MSSD), and Maximum Symmetry-Aware Projection Distance (MSPD). The final average recall (AR) is calculated by averaging the individual average recall scores of these three metrics across a range of error thresholds. For a more detailed explanation, refer to the BOP challenge~\cite{hodan2023bop}.

\begin{table*}[ht]
\centering
\small
\setlength{\tabcolsep}{0.6mm}

\begin{tabular}{cccc|cccccccc|c}
\hline
Method & Parm size  & Ref Num & Detection  & LM-O           & T-LESS         & TUD-L          & IC-BIN       & ITODD         & HB            & YCB-V          & MEAN  & TIME              \\ \hline
OSOP~\cite{shugurov2022osop} & -     & 90 k     & OSOP~\cite{shugurov2022osop}      & 27.4          & -          & -             & -           & -             & 46.4             & 29.6          & -                & - \\
MegaPose~\cite{labbe2022megapose} & 21.6 M & 520     & Mask R-CNN~\cite{he2017mask} & 18.7          & 19.7          & 20.5          & 15.3        & 8.0           & 18.6          & 13.9          & 16.4             & - \\ 
\hline
ZS6D~\cite{ausserlechner2023zs6d} & 21.7 M    & 300     & CNOS~\cite{nguyen2023cnos}       & 29.8          & 21.0          & -             & -           & -             & -             & 32.4          & -               & -  \\
MegaPose~\cite{labbe2022megapose} & \uline{21.6} M & 520     & CNOS~\cite{nguyen2023cnos}       & 22.9          & 17.7          & 25.8          & 15.2        & 10.8          & 25.1          & 28.1          & 20.8            & 15.5 s  \\
GigaPose~\cite{nguyen2024gigaPose} & 316.3 M& 162     & CNOS~\cite{nguyen2023cnos}      & 29.9 & 27.3 & 30.2          & 23.1        & \uline{18.8} & 34.8          & 29.0          & 27.6            & 0.8 s  \\
Genflow~\cite{moon2024genflow} & 21.7 M& 208     & CNOS~\cite{nguyen2023cnos}      & 25.0 & 21.5 & 30.0          & 16.8        & 15.4 & 28.3          & 27.7          & 23.5            & 3.8 s\footnotemark[1] \\
FoundPose~\cite{ornek2023foundpose} & 302.9 M   & 798      & CNOS~\cite{nguyen2023cnos}    & \textbf{39.6}         & \textbf{33.8}  & \textbf{46.7} & 23.9 & \textbf{20.4}  & \textbf{50.8} & 45.2 & \textbf{37.2} & 1.6 s \\
MixRI (ours)   & \textbf{5.3 M}  & \textbf{12}      & CNOS~\cite{nguyen2023cnos}      & 27.0          & 25.4          & 29.3 & \uline{29.7} & 10.9          & 44.9 & \uline{52.8} & 31.4 & \textbf{0.5 s} \\ 
MixRI (ours) & \textbf{5.3 M}   & \uline{24}      & CNOS~\cite{nguyen2023cnos}      & \uline{30.4}          & \uline{27.4}          & \uline{33.6} & \textbf{30.8} & 11.6          & \uline{50.2} & \textbf{54.6} & \uline{34.1} & \uline{0.7 s}  \\
\hline
	
\end{tabular}
\vspace{-0.5em}
\caption{\textbf{Results on the seven core BOP datasets.} The table compares methods in Average Recall (AR), network parameter size, reference image count, and inference time. \textbf{Bold} denotes the best, and \uline{underline} the second best.} 
\label{tab:main}
\vspace{-1.em}
\end{table*}
\begin{table}[t]
\small
\centering
\setlength{\tabcolsep}{0.1cm}
\begin{tabular}{cccccc}
\hline
Ref Nums      & Method     & YCB-V         & LM-O          & TUD-L         & MEAN          \\ \hline
\multirow{2}{*}{4}  & GigaPose~\cite{nguyen2024gigaPose}   & 8.1           & 10            & 9.1           & 9.1           \\ \cline{2-6} 
                    & MixRI (ours) & \textbf{28.6} & \textbf{11.2} & \textbf{13.8} & \textbf{17.9} \\ \hline
\multirow{2}{*}{6}  & GigaPose~\cite{nguyen2024gigaPose}   & 7.9           & 12            & 13.1          & 11.0          \\ \cline{2-6} 
                    & MixRI (ours) & \textbf{46.5} & \textbf{22.1} & \textbf{25.7} & \textbf{31.4} \\ \hline
\multirow{2}{*}{8}  & GigaPose~\cite{nguyen2024gigaPose}   & 15.4          & 15.6          & 12.3          & 14.4          \\ \cline{2-6} 
                    & MixRI (ours) & \textbf{51.8} & \textbf{26.2} & \textbf{28.5} & \textbf{35.5} \\ \hline
\multirow{2}{*}{12} & GigaPose~\cite{nguyen2024gigaPose}   & 14.9          & 13.9          & 15.3          & 14.7          \\ \cline{2-6} 
                    & MixRI (ours) & \textbf{52.8} & \textbf{27.0} & \textbf{29.3} & \textbf{36.4} \\ \hline
\multirow{2}{*}{24} & GigaPose~\cite{nguyen2024gigaPose}   & 15.9          & 18.2          & 20.8          & 18.3          \\ \cline{2-6} 
                    & MixRI (ours) & \textbf{54.6} & \textbf{30.4} & \textbf{33.6} & \textbf{39.5} \\ \hline
\end{tabular}
\caption{\textbf{Comparison with a limited number of reference images.} We compare our method with GigaPose in different number of reference images.}
\label{tab:refnums}
\vspace{-1.5em}
\end{table}
\subsection{Comparison with the State of the Art}
\begin{figure}[t]
    \centering
    \includegraphics[width=0.48\textwidth]{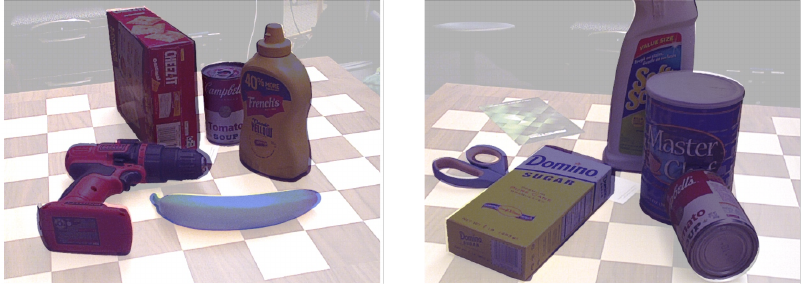}
    \caption{\textbf{Qualitative results on YCB-V.} We present the pose estimation results obtained using MixRI. All the results are visualized in error heatmap~\cite{tremblay2023diffdope} which darker blue indicates lower error with respect to the ground truth pose (legend: 0 cm \includegraphics[height=0.2cm] {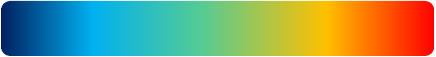} 5 cm).}
    \label{fig:qualitative}
    \vspace{-2em}
\end{figure}

We compare our method with OSOP~\cite{shugurov2022osop}, MegaPose~\cite{labbe2022megapose}, ZS6D~\cite{ausserlechner2023zs6d}, GigaPose~\cite{nguyen2024gigaPose}, Genflow \cite{moon2024genflow} and  FoundPose \cite{ornek2023foundpose} as shown in \Cref{tab:main}. In a setting without refinement, our method achieves promising results. Specifically, compared to FoundPose, our method achieves similar performance while utilizing 33$\times$ fewer reference images, with fewer network parameters and shorter inference time. When compared to GigaPose, which uses the second fewest reference images, our method improves accuracy by approximately 10\% on the challenging IC-BIN and HB datasets, and by around 25\% on the challenging YCB-V dataset. We also achieve comparable results on LM-O, T-LESS, and TUD-L. However, the performance on ITODD is not as good as GigaPose. This might be because it is a dataset with grayscale images, whereas our training is based on RGB images. In addition, it has severe occlusions, reflections, and surfaces with weak textures, which increases the difficulty of matching. However, we still improve the mean AR of all seven core datasets by 6.5\% compared with GigaPose. When it comes to MegaPose, our method achieves better results across all seven core datasets and improves around 15\% on average. Genflow utilizes a network similar to MegaPose, and our method achieves better results on most datasets, except for ITODD, where performance is impacted due to differences in color modes. ZS6D is a method that first retrieves the closest reference image from 300 reference images and matches the query image with the closest one. However, our method uses far fewer reference images but achieves better results. This demonstrates that our multi-view fusion strategy can improve matching accuracy and increase the AR score. \Cref{fig:qualitative} shows qualitative results, which illustrate the accurate pose estimation results. More qualitative results and comparisons with other methods are provided in the supplementary material.
\footnotetext[1]{Since GenFlow's code is not open-source, the reported time here is taken from its public report in BOP, using a V100 GPU.}

\noindent\textbf{Limited Number of Reference Images.} Since GigaPose uses the fewest reference images among existing works, we compare it when using only a limited number of reference images. We conduct experiments with various settings by changing the reference image number $S$ and report the results in \Cref{tab:refnums}. GigaPose experiences a significant performance decrease when there are fewer reference images. Because it's a two-stage approach, which highly relies on the closest reference image. When there are few reference images, selecting a reference image with similar viewpoint becomes difficult, and it is also challenging to provide sufficient information for estimating the pose in their second stage. It is worth noting that other methods \cite{labbe2022megapose, ornek2023foundpose, moon2024genflow, ausserlechner2023zs6d}, which are also based on the retrieval mechanism, can encounter the similar issue when using a limited number of reference images. In contrast, our method is designed specifically for limited reference images and achieve a better AR score when using the same number of reference images.
\begin{table}[b]
\small
\vspace{-1.em}
\centering
\begin{tabular}{cc|cccc}
\hline
SAP & MARQ & YCB-V & LM-O & TUD-L & MEAN \\ \hline
\color{red} \ding{55}          & \color{red} \ding{55}            & 2.9   & 2.8  & 2.2   & 2.6  \\ 
\color{green} \ding{51}         & \color{red} \ding{55}            & 19.2  & 7.8  & 8.7   & 11.9 \\ 
\color{red} \ding{55}          & \color{green} \ding{51}           & 33.2  & 13.4 & 21.9  & 22.8 \\ \hline
\color{green} \ding{51}         & \color{green} \ding{51}           & \textbf{54.6}  & \textbf{30.4} & \textbf{33.6}  & \textbf{39.5} \\ \hline
\end{tabular}
\caption{\textbf{Effectiveness of the attention module.} We analyze the impact of our different modules. Here SAP stands for Self-Attention between Points in~\Cref{attention} and MARQ stands for Mix-Attention between Reference \& Query in~\Cref{attention}.}
\label{tab:ablation}
\vspace{-1.em}
\end{table}

\noindent\textbf{Run-time \& Memory.} 
We report the average inference timings for each method in \Cref{tab:main}, measured using a single 4090 GPU. It is worth noting that the detection time included in CNOS \cite{nguyen2023cnos} is also taken into account, and each image may contain multiple objects for pose estimation. In addition to reducing inference time, our method significantly saves time in rendering the reference images and does not require pre-extraction of features. However, GigaPose and FoundPose need pre-extracting all the reference image features \cite{nguyen2024gigaPose,ornek2023foundpose}, which also waste memory due to the need to cache them. Compared to MegaPose, which also does not require pre-extracting features, our method outperforms it, as we eliminate the need for image rendering during inference \cite{labbe2022megapose}. 

We also report the network parameters compared with others in \Cref{tab:main}. The parameter count of our network is significantly smaller than that of all other methods. In particular, we have over $50\times$ fewer parameters than GigaPose. Combined with fewer reference images and no feature cache needed, MixRI is very friendly for deployment on devices with limited memory space. More discussions are shown in \ref{sec:extraction} in the supplementary material.

\subsection{Ablation Study}
\label{sec:ablation}
\textbf{Effectiveness of the Attention Module.} Our main innovation is to fuse the multiple reference points in the feature space. To demonstrate the effectiveness of the fusing strategy, we conduct several ablation studies. As shown in \Cref{tab:ablation}, through row 1 and row 2, adding SAP brings a 9.3\% increase in AR score. The considerable increase comes with MARQ, which results in a 20.2\% increase in AR score. Finally, with both modules, we get the final AR score shown in row 4, which has a 36.9\% increase compared to row 1.

\begin{figure}[t]
    \centering
    
    \vspace{-0.8em}
    \includegraphics[width=0.48\textwidth]{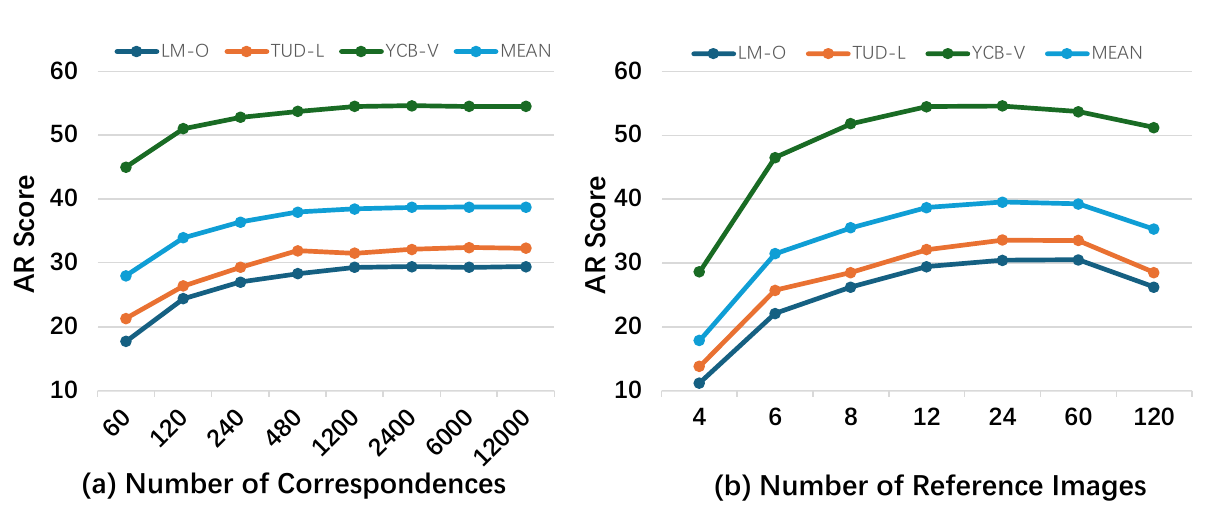}
    \caption{\textbf{Impact of the number of correspondences and reference images.} We report the AR score on LM-O, TUD-L, and YCB-V, along with the average.}
    \label{fig:cor_ref_nums}
    \vspace{-1em}
\end{figure}

\noindent\textbf{Number of Correspondences.} \Cref{fig:cor_ref_nums} (a) shows the results of different correspondences used in matching. Our method is robust to correspondence, and increasing the correspondences can increase the AR score until around 2400. It is worth mentioning that, although there are only 60 correspondences, for the YCB-V dataset, there is already an AR score of 45\%. When the number of correspondences reaches 240, it is already comparable to other methods.

\begin{table*}[t]
\centering
\small
\setlength{\tabcolsep}{0.6mm}
\begin{tabular}{cccc|cccccccc|c}
\hline
Method & Ref Num & Detection &Refinement  & LM-O           & T-LESS         & TUD-L          & IC-BIN       & ITODD         & HB            & YCB-V          & MEAN & TIME      \\ \hline
MixRI+ (ours)  & 162      & CNOS~\cite{nguyen2023cnos}  & -    &  42.6       & 36.0  & 41.8 & 33.9 & 22.2 & 56.0 & 56.9 & 41.3 & 1.0 s \\
\hline
MegaPose~\cite{labbe2022megapose}   & 520      & CNOS~\cite{nguyen2023cnos}  & MegaPose~\cite{labbe2022megapose}    & 49.9          & 47.7          & \textbf{65.3} & 36.7 & 31.5          & 65.4 & 60.1 & 50.9 & 17.0 s \\
GigaPose~\cite{nguyen2024gigaPose}  & 162      & CNOS~\cite{nguyen2023cnos}  & MegaPose~\cite{labbe2022megapose}    & \textbf{55.6}          & \textbf{54.6}          & 57.8 & 44.3 & \textbf{37.8}          & \uline{69.3} & 63.4 & \textbf{54.7} & 2.3 s \\
FoundPose~\cite{ornek2023foundpose}   & 798      & CNOS~\cite{nguyen2023cnos}  & MegaPose~\cite{labbe2022megapose}    & \uline{55.4}          & \uline{51.0}          & \uline{63.3} & 43.0 & \uline{34.6}          & \textbf{69.5} & \textbf{66.1} & \textbf{54.7} & 4.4 s\\
MixRI (ours)   & \textbf{12}      & CNOS~\cite{nguyen2023cnos}  & MegaPose~\cite{labbe2022megapose}    &  40.8       & 45.0  & 44.6 & 44.4 & 21.6 & 61.4 & 61.1 & 45.6 & \textbf{1.5 s} \\
MixRI (ours)   & \uline{24}      & CNOS~\cite{nguyen2023cnos}  & MegaPose~\cite{labbe2022megapose}    &           44.8 &46.1 & 48.2 & \textbf{44.6} & 21.3 & 61.4 & 62.0 & 46.9 & \uline{1.7 s}\\
MixRI+ (ours)  & 162      & CNOS~\cite{nguyen2023cnos}  & MegaPose~\cite{labbe2022megapose}    &           52.1 &50.4 & 57.5 & \uline{44.5} & 34.3 & 67.1 & \uline{64.8} & \uline{53.0} & 2.1 s \\
\hline
	
\end{tabular}
\caption{\textbf{Results on the seven core BOP datasets with refinement.} The table shows Average Recall (AR) scores per dataset, the time and the number of reference images. We also present MixRI+, the same structure but trained for more reference images available.}
\label{tab:refinement}
\vspace{-0.5em}
\end{table*}

\noindent\textbf{Number of Reference Images.} \Cref{fig:cor_ref_nums} (b) ablates the number of reference images. Our network is trained using 12 reference images, but it does not require the same number of reference images during inference. Therefore, we only alter the number of reference images during the test stage. However, we observe a performance degradation when the number of reference images increases significantly, particularly when it exceeds 60. This is expected because our lightweight network is less capable of handling the feature fusion from a large number of reference images. As previous work \cite{labbe2022megapose, nguyen2024gigaPose,ornek2023foundpose}, more reference images need a larger network with more parameters to handle.

\noindent\textbf{Effectiveness of Occlusion Flags.}
\label{sec:occlusion_flag}
\Cref{tab:occlusion_flag} demonstrates the effectiveness of incorporating occlusion flags. It is worth noting that when $\tau_\text{occ} = 1$, the occlusion flag mechanism is effectively disabled. The results validate the robustness of the method when occlusion prediction is enabled.
\begin{table}[t]
    \centering
    \small
    \setlength{\tabcolsep}{1.3pt}
    \begin{tabular}{c|cccccccc}
    \hline
    $\tau_\text{occ}$ & LM-O & T-LESS & TUD-L & IC-BIN & ITODD & HB & YCB-V & MEAN  \\ \hline
    1.0 & 27.6 & 24.1 & 30.9 & 28.5 & 10.4 & 44.1 & 49.4 & 30.7 \\ 
    0.8 & \textbf{30.4}   & \textbf{27.4} & \textbf{33.6} & \textbf{30.8} & \textbf{11.6} & \textbf{50.2} & \textbf{54.6} & \textbf{34.1} \\ 
    0.5 & 27.8 & 26.7 & 28.4 & 28.6 & 10.2 & 46.8 & 53.8 & 31.8\\ \hline
    \end{tabular}
    \caption{\textbf{The effectiveness of occlusion flags.} We compare different occlusion flag settings.}
    \label{tab:occlusion_flag}
    \vspace{-1.5em}
\end{table}

\noindent\textbf{Refinement.} While refinement can lead to improved pose accuracy, it does so at the expense of speed, which goes against our initial intention. However, to demonstrate the potential of our lightweight network, we also present the results after refinement in \Cref{tab:refinement}. For this, we use MegaPose \cite{labbe2022megapose} to further refine our initial pose estimates. Additionally, we include another MixRI variant (MixRI+) having the same structure but trained for more reference images available. To optimize the use of reference images, we follow the approach in \cite{nguyen2024gigaPose}, which retrieves the reference image with the closest out-of-plane rotation. However, in our method, we also select three additional reference images as input to the network. Detailed information on this process is provided in the supplementary material. Notably, with more reference images available, the initial poses achieve an AR score of 41.3, surpassing FoundPose which is 37.2. Additionally, the refinement performance is comparable, even we use fewer network parameters and achieve faster processing speed.
\vspace{-0.4em}
\section{Conclusion}
\label{sec:conclusion}
\vspace{-0.4em}
In this paper, we proposed MixRI, a novel approach for CAD-based novel object pose estimation using RGB images. Designed with practical applications in mind, our method features a lightweight network and requires far fewer reference images than existing approaches, eliminating the need for feature caching. This reduces both pose inference time and reference image preparation time while avoiding pre-processing. Despite its simplicity, MixRI achieves performance comparable with more complex networks that rely on numerous reference images. In the future, we will further improve the matching accuracy and extend our framework to scenarios where object models are unavailable.

{\small
\bibliographystyle{ieeenat_fullname}
\bibliography{11_references}
}

\ifarxiv \clearpage \appendix \section{Appendix}
\label{sec:appendix_subsection}

In this section, we provide detailed information about our work. In \Cref{sec:ground truth-2d-2ds}, we describe the procedure for generating the ground truth correspondences. In \Cref{sec:referece_pool}, we outline the strategy for selecting reference images in our experiments. In \Cref{sec:variant}, we present the design of MixRI+, the variant of MixRI. Next, we discuss some limitations and failure cases in \Cref{sec::limit} and detail the evaluation datasets in \Cref{sec:evaluation datasets}. We also provide qualitative results in \Cref{sec:qualitative}. In \Cref{sec:extraction} we discuss the balance between time and memory cost in different methods. Finally, we detail the training process in \Cref{sec:training-details} and address ethical considerations in \Cref{sec::Ethics}.

\subsection{Ground Truth Correspondences}
\label{sec:ground truth-2d-2ds}
\begin{figure*}[b]
    \centering
    \includegraphics[width=0.95\textwidth]{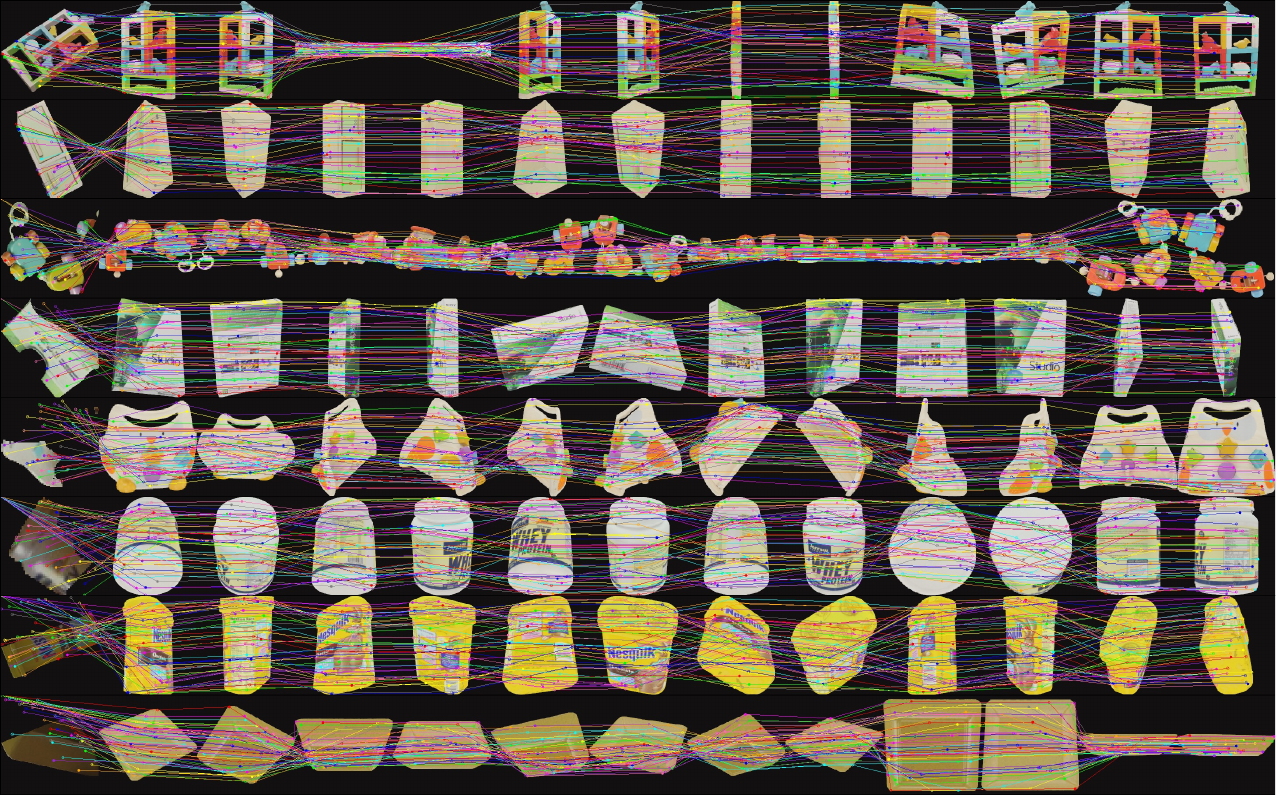}
    \caption{\textbf{Training samples.} Each row represents one batch. The left-most image is the query image and the remaining are 12 reference images. We show the correspondences across all reference images with the query image. Solid points represent visible points, while hollow points indicate occluded points. For the query image, some points are projected outside of the image. We set them to the location of $[-1,-1]$ and regard them as occluded points during training.}
    \label{fig:training sample}
\end{figure*}

We train our network entirely with synthetic images, using the ground truth 3D information provided in GSO-Datasets~\cite{labbe2022megapose}. To sample the 3D object points, we sample the 2D image points in all reference images and unproject the 2D image points into the 3D space. For each reference image, we randomly select $M$ image points within the object mask and find their corresponding points along all the other reference images with the query image using the ground truth pose and rendered depth. We also use depth to judge whether the corresponding location is occluded and set $\tau$ to 4mm. In our default setting, $M$ is 10, and $S$ is 24, which produces $N = 240$ points in total.

Because depth can have a sharp change in edge, we use the close operation in morphology to shrink the mask. We set the kernel size to $3 \times 3$ and repeat 3 iterations. If the mask is really small and cannot sample enough points, we just repeat the sampled points.

We use the same procedures during the evaluation to produce the correspondences along all the reference images as training procedures, except for the query image.

We also use some augmentations during training, such as Gaussian blur, contrast, brightness, sharpness, and color change, as done in \cite{nguyen2024gigaPose}. We provide some illustrations to show the training samples in \Cref{fig:training sample}.

\begin{figure*}[b]
    \centering
    \includegraphics[width=\textwidth]{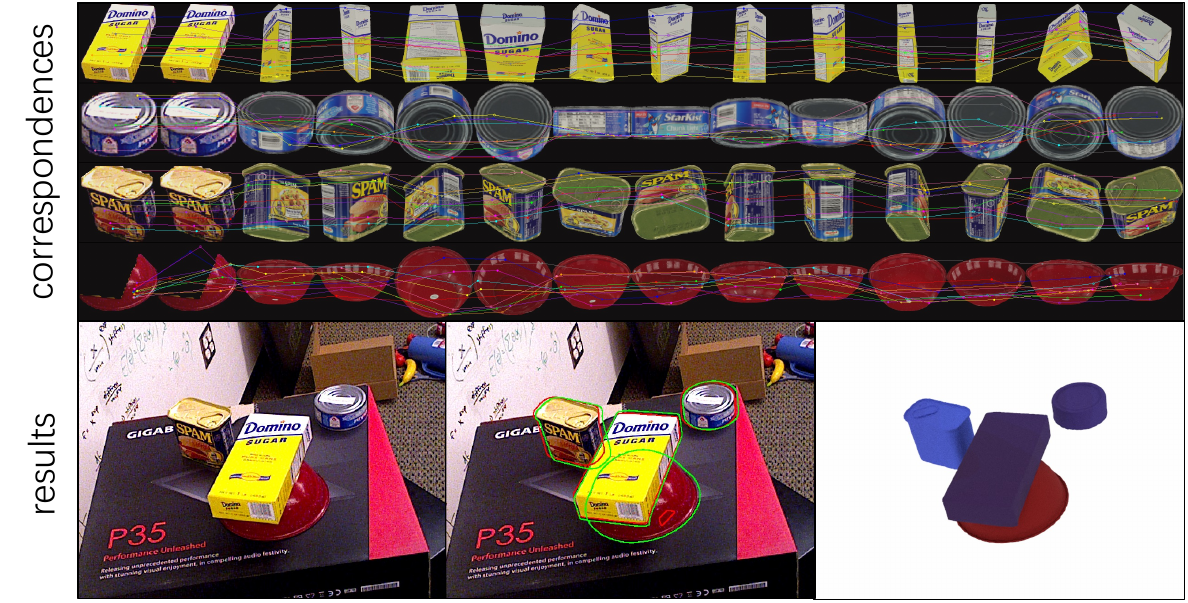}
    \caption{\textbf{Failure Case.} The bowl is texture-less and nearly occluded, which make the matching really challenge. The top shows the matching results, where the point on the far left image is the predicted matched point, the second image from the left shows the ground truth matched point, and the remaining images are reference images, totaling 12. So if the lines between the first and second images on the left are parallel and of equal length, it means the match is correct. For ease of viewing, we randomly sampled 10 points predicted to be visible. The lower part of each result set displays the estimated results. The far-left image is the RGB image, the middle image shows the projection of the ground truth pose (in \textcolor{green}{green}) and the estimated pose (in \textcolor{red}{red}). The image on the far right displays the error heatmap calculated between the ground truth pose and the predicted pose which darker red indicates higher error with respect to the ground truth pose (legend: 0 cm
    \includegraphics[height=0.2cm] {legend.pdf} 5 cm).}
    \label{fig:failure}
\end{figure*}

\subsection{Reference Images Selection Strategy}
\label{sec:referece_pool}
In the default setting, our method uses only 24 reference images to have a balance of accuracy and performance. To demonstrate the generality of our method, all reference images are randomly selected from a reference image bank for each inference. In detail, to generate the reference image bank, we generate 162 reference images from viewpoints defined on a regular icosphere, which is created by subdividing each triangle of the Blender icosphere primitive into four smaller triangles, just as \cite{nguyen2024gigaPose} does. We use the farthest sampling strategy (FPS) during each inference session to sample $S$ (in most experiments, $S=24$) reference images. Because in-plane rotations do not provide additional information about the visibility of 3D object points, we only measure the out-of-plane distance between two rotations:
\begin{equation}
    d\left(\hat{\mathbf{R}}_{0}, \hat{\mathbf{R}}_{1}\right)=\arccos \left(\frac{\operatorname{tr}\left(\hat{\mathbf{R}}_{0}^{\top} \hat{\mathbf{R}}_1\right)-1}{2}\right) / \pi,
\end{equation}

where $\hat{\mathbf{R}}_i$ means rotation removed the in-plane components from the origin rotation matrix $\mathbf{R}_i$. It is worth mentioning that for real-world scenarios, we can omit the process of establishing the reference image bank and directly generate only $S$ reference images, pre-sampling the corresponding relationships between them in advance, as stated in \Cref{sec:ground truth-2d-2ds}.

\subsection{MixRI Variant}
\label{sec:variant}
MixRI is designed for faster inference, a smaller network cache, and faster preparation of reference images. However, we also consider scenarios where more reference images are available, allowing for further accuracy improvements. To effectively select the most relevant reference images from a large reference image pool, we follow other approaches \cite{nguyen2024gigaPose, ornek2023foundpose} to identify the most suitable reference images. Here we use the same reference image bank as stated in \Cref{sec:referece_pool} which has 162 reference images in total. We use the feature extractor from \cite{nguyen2024gigaPose} as our feature extractor, as it also mitigates the impact of in-plane rotations, similar to our framework. Following \cite{nguyen2024gigaPose}, we identify the top-1 candidate with the most similar out-of-plane rotations and then select three additional reference images based on their rotation alignment with the retrieved reference image. To effectively gather the closest viewpoint information, we train a MixRI variant which takes only four relevant reference images but keeps the model structure, model size the same. This variant is designed for situations where more reference images are available and the reference images have closer relationships to each other.

\begin{figure}[b]
    \centering
    \includegraphics[width=0.48\textwidth]{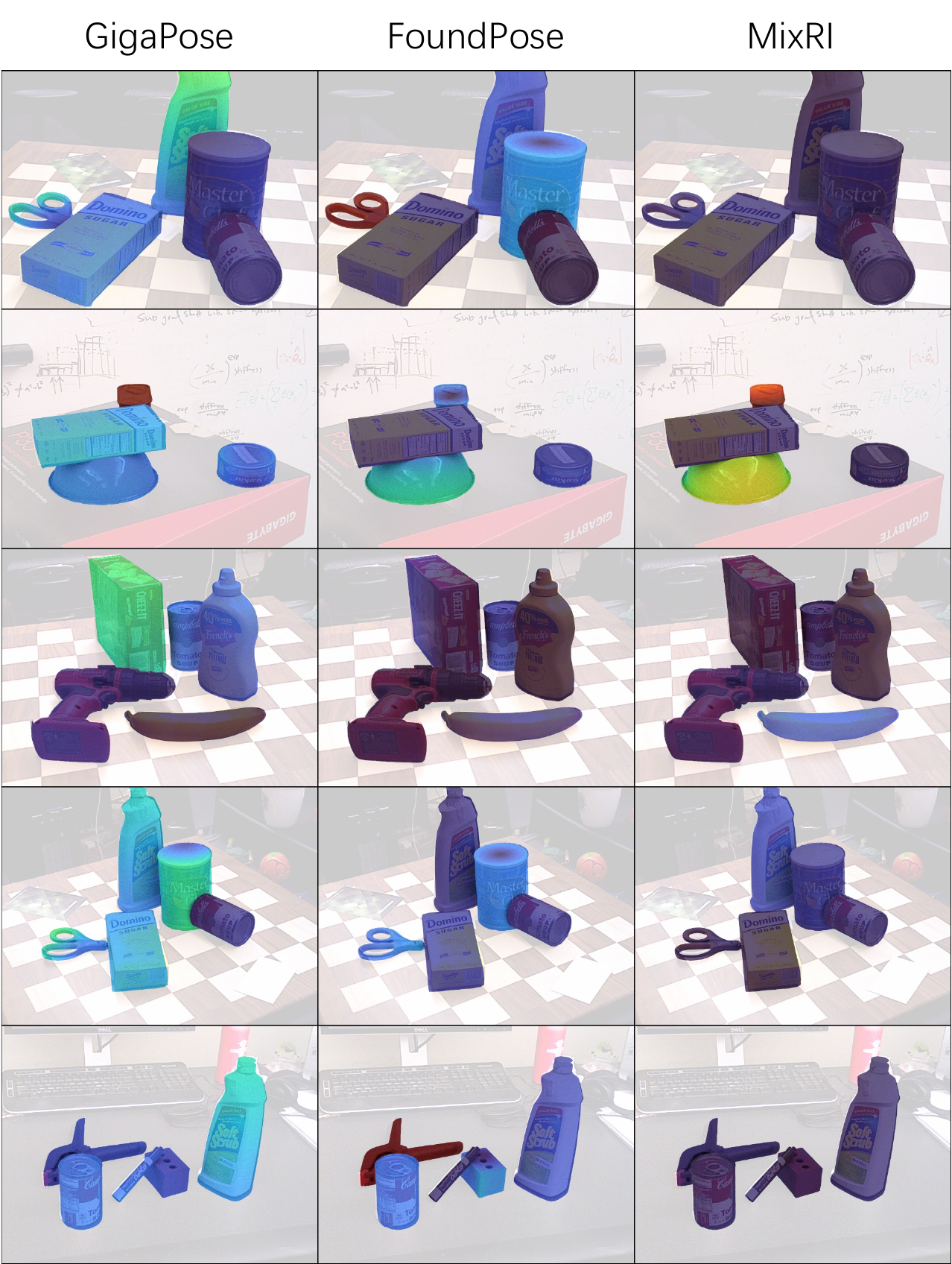}
    \caption{\textbf{Qualitative comparison with GigaPose \cite{nguyen2024gigaPose} and FoundPose \cite{ornek2023foundpose}}.We compare our method with GigaPose and FoundPose, which are two leading methods in speed and accuracy. All the results are visualized in error heatmap calculated between the ground truth pose and the predicted pose which darker red indicates higher error with respect to the ground truth pose (legend: 0 cm
    \includegraphics[height=0.2cm] {legend.pdf} 5 cm).}
    \label{fig:vis-compare}
\end{figure}
\subsection{Limitations and Failure Cases}
\label{sec::limit}
The limitations of our method stem from classic challenges in matching, such as matching weak textures and similar areas. Our method may fail when applied to highly textureless objects. Although we try to handle such situations using attention mechanisms, our method can struggle to match between similar patches, especially when the detection result only captures a part of the whole object or object that is significantly occluded. As shown with the bowl in Figure \ref{fig:failure}, it is almost completely obscured and with very weak texture. As the visualization shows, the points considered visible in the query image actually mostly lie in the occluded areas, leading to an incorrect pose estimation in the end.

\subsection{Evaluation Datasets}
\label{sec:evaluation datasets}
We evaluate our method on seven core BOP datasets~\cite{hodan2023bop}, including LM-O~\cite{brachmann2014learning}, YCB-V~\cite{xiang2017posecnn}, T-LESS~\cite{hodan2017tless}, TUD-L~\cite{hodan2018bop}, IC-BIN~\cite{doumanoglou2016recovering}, HomebrewedDB(HB)~\cite{kaskman2019homebreweddb} and ITODD~\cite{drost2017introducing}. These datasets have 132 objects in total, which are never seen during the training stage. They include the general challenge of the RGB pose estimation, such as illumination change, texture-less object, strong occlusion, and cluttered scenes. 

\subsection{Qualitative Results}
\label{sec:qualitative}
In \Cref{fig:vis-compare}, we present qualitative comparisons with GigaPose \cite{nguyen2024gigaPose} and FoundPose \cite{ornek2023foundpose} , which are the two leading methods in speed and accuracy, respectively. Our method can make some mistakes especially in texture-less objects and some really small objects. However, we still achieve comparable results and can have better performance in some cases.

We illustrate more qualitative examples of our method in \Cref{fig:vis-ycbv} for YCB-V \cite{xiang2017posecnn} and in \Cref{fig:vis-lmo} for LM-O \cite{brachmann2014learning}. All illustrations are composed of the original RGB image, the corresponding matching results, the projection of the object model with ground truth pose and predicated pose, and the error heatmap.



\subsection{Time \& Memory Balance}
\label{sec:extraction}
 In \Cref{tab:extraction}, we compare the feature extraction costs for a single object on an RTX 4090 GPU. Both GigaPose \cite{nguyen2024gigaPose} and FoundPose \cite{ornek2023foundpose} require significant time for feature extraction and pre-processing. On one hand, they rely on a large number of reference images, increasing the time required to extract features from all reference images. On the other hand, they include a view selection stage to identify the closest reference image, introducing additional pre-processing overhead, such as clustering. This time-consuming process follows a space-for-time strategy---pre-processing during the inference stage and caching them, thus leading to increased memory consumption. For memory-limited devices, this constraint reduces the number of objects whose poses can be estimated, as additional memory is required to store reference images, pre-extracted features, and large network parameters. In contrast, MixRIs are highly efficient. Thanks to the fewer reference images, we omit pre-extraction and instead perform feature extraction during inference, significantly reducing memory requirements with no cache needed. With fewer reference images and a lightweight network, MixRIs are more suitable for memory-constrained devices but still keep fast inference speed. 

\begin{table}[h]
\small
\setlength{\tabcolsep}{1.3pt}
\begin{tabular*}{0.48\textwidth}{@{\extracolsep{\fill}}c|cccc}
\hline
Method & GigaPose \cite{nguyen2024gigaPose} & FoundPose \cite{ornek2023foundpose} & MixRI(12) & MixRI(24) \\ \hline
Time   & 11.6 s   & 40 s     & \textbf{21 ms} & 42 ms      \\
Memory & 233.5 M  & 523.5 M  & \textbf{9.2 M} & 18.4 M     \\ \hline
\end{tabular*}
\caption{\textbf{The time and memory cost for processing all the reference images of one object}. MixRIs are much more efficient than GigaPose and FoundPose, making it more suitable for real-world applications.}
\label{tab:extraction}
\end{table}

\subsection{Training Details}
\label{sec:training-details}
We train our network from scratch. During training, we use the GSO-Datasets provided by MegaPose~\cite{labbe2022megapose}, which includes over 1 million images generated by BlenderProc~\cite{Denninger2023}. We train our networks using PyTorch with the AdamW optimizer with $\beta =(0.9, 0.999)$  and 5e-4 weight decay. The learning rate is set to 1e-4 and the warm up is 200 iterations. We train our network around 600k iterations, which spend about one week on four 4090 GPUs with batch size 8 on each. We use 12 reference images to train our network and $N$ is set to 240. All images are cropped to $224 \times 224$. We also add some augmentations as done in~\cite{nguyen2024gigaPose} like: Gaussian blur, contrast, brightness, sharpness, and color change. The feature dimension $D$ is set to 256. The number of iterations for each module 
$n_0,n_1,n_2,n_3$ are set to 4, 2, 2, 2, respectively. Besides, for each module, the weights are shared during the iteration. We set $\tau_\text{occ} = 0.8$.

For the training losses, the predicted 2D coordinates are normalized to the range $[-1, 1]$ for stable training. The Huber delta is set to 0.0357, corresponding to 4 pixels before normalization. 

\subsection{Ethics}
\label{sec::Ethics}
This research contributes to the development of object pose estimation, with potential applications in robotics, augmented reality (AR), and machine vision in general. While many of those applications could bring societal benefits (e.g. workload decrease through automation, AR-based teaching or assistance), it could also be used for unethical purposes.

\begin{figure*}[htbp]
    \centering
    \includegraphics[width=\textwidth]{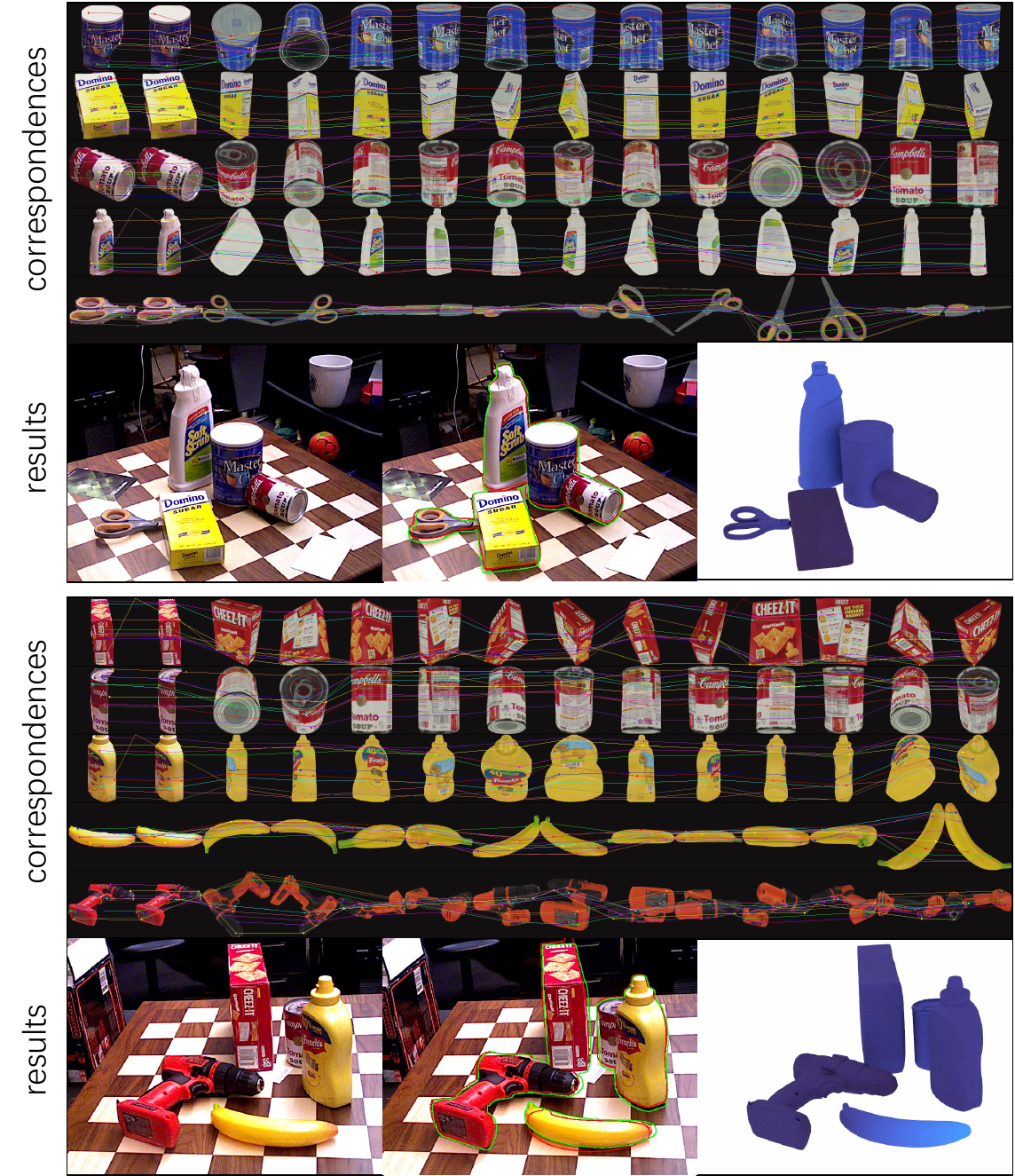}
    \caption{\textbf{Qualitative results on YCB-V \cite{xiang2017posecnn}}. We presented two sets of results, with the top of each set showing the matching results. In these, the point on the far left image is the predicted matched point, the second image from the left shows the ground truth matched point, and the remaining images are reference images, totaling 12. So if the lines between the first and second images on the left are parallel and of equal length, it means the match is correct. For ease of viewing, we randomly sampled 10 points predicted to be visible. The lower part of each result set displays the estimated results. The far-left image is an RGB image, the middle image shows the projection of the ground truth pose (in \textcolor{green}{green}) and the estimated pose (in \textcolor{red}{red}). The image on the far right displays the error heatmap calculated between the ground truth pose and the predicted pose which darker red indicates higher error with respect to the ground truth pose (legend: 0 cm
    \includegraphics[height=0.2cm] {legend.pdf} 5 cm).}
    \label{fig:vis-ycbv}
\end{figure*}

\begin{figure*}[ht]
    \centering
    \includegraphics[width=\textwidth]{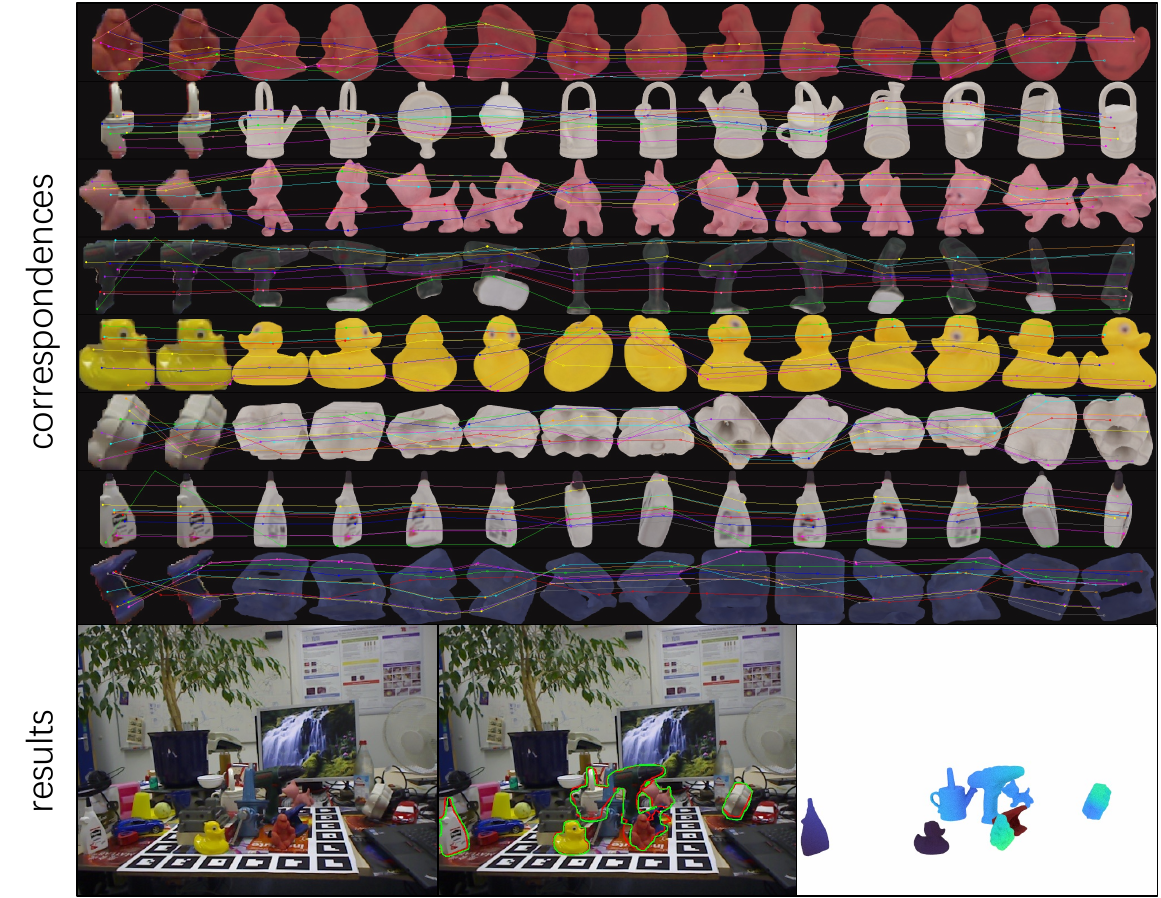}
    \caption{\textbf{Qualitative results on LM-O \cite{brachmann2014learning}}. The top shows the matching results. In these, the point on the far left image is the predicted matched point, the second image from the left shows the ground truth matched point, and the remaining images are reference images, totaling 12. So if the lines between the first and second images on the left are parallel and of equal length, it means the match is correct. For ease of viewing, we randomly sampled 10 points predicted to be visible. The lower part of each result set displays the estimated results. The far-left image is an RGB image, the middle image shows the projection of the ground truth pose (in \textcolor{green}{green}) and the estimated pose (in \textcolor{red}{red}). The image on the far right displays the error heatmap calculated between the ground truth pose and the predicted pose which darker red indicates higher error with respect to the ground truth pose (legend: 0 cm
    \includegraphics[height=0.2cm] {legend.pdf} 5 cm).}
    \label{fig:vis-lmo}
\end{figure*}
\newpage \fi

\end{document}